\documentclass[journal,twoside,web]{ieeecolor}
\usepackage{generic}
\usepackage{cite}
\usepackage{amsmath,amssymb,amsfonts}
\usepackage{threeparttable}
\usepackage{algorithmic}
\usepackage{graphicx}
\usepackage{textcomp}
\usepackage{balance}
\usepackage[subrefformat=parens,labelformat=parens]{subfig}
\usepackage{multirow}
\usepackage{bbm}
\usepackage{soul}
\usepackage{mathtools}
\usepackage{pifont}
\newcommand{\cmark}{\ding{51}}%
\newcommand{\xmark}{\ding{55}}%

\graphicspath{ {figures/} }

\def\spo2{SpO$_2$}
\def\R{\mathbb{R}}

\newcommand{\rev}[1]{{\color[RGB]{0,0,0}#1}}

\ifodd 1

\newcommand{\CommentWong}[1]{\textcolor[rgb]{1,0,0}{[Wong: #1]}}

\else

\newcommand{\CommentWong}[1]{}

\fi

\begin{document}

\title{Remote Blood Oxygen Estimation From Videos Using Neural Networks}

\author{Joshua~Mathew*,~\IEEEmembership{Graduate Student Member,~IEEE,}
Xin~Tian*,~\IEEEmembership{Graduate Student Member,~IEEE,} Chau-Wai~Wong,~\IEEEmembership{Member,~IEEE,}
{Simon~Ho, Donald~K.~Milton},
and~Min~Wu,~\IEEEmembership{Fellow,~IEEE}
\thanks{* J.~Mathew and X.~Tian equally contributed to this paper.}
\thanks{J.~Mathew was with the Department of Electrical and Computer Engineering, NC State University, Raleigh, NC, 27695 USA (email: joshuaregimathew@gmail.com).}
\thanks{C.-W. Wong is with the Department of Electrical and Computer Engineering, Forensic Sciences Cluster, and Secure Computing Institute, NC State University, Raleigh, NC, 27695 USA (email: chauwai.wong@ncsu.edu).}
\thanks{X. Tian was with the Department of Electrical and Computer Engineering, University of Maryland, College Park, MD, 20742 USA (e-mail: xtian17@umd.edu).}
\thanks{M. Wu is with the Department of Electrical and Computer Engineering, University of Maryland, College Park, MD, 20742 USA (e-mail: minwu@umd.edu).}
\thanks{{S.~Ho is with the Physical Therapy and Rehabilitation Science, University of Maryland School of Medicine, Baltimore, MD, 21201 USA {(email: simon.ho@som.umaryland.edu)}.}}
\thanks{{D.~K.~Milton is with the Maryland Institute
for Applied Environmental Health,
School of Public Health, University of Maryland, College Park, MD 20742, USA (e-mail: dmilton@umd.edu).}}
\thanks{This work was supported {in part} by NSF ECCS under Grants 2030502, 2030430, and 2030382, a UM Venture COVID Challenge award, the National Institute of Allergy and Infectious Diseases (NIAID) Centers of Excellence for Influenza Research and Surveillance (contract number HHSN272201400008C), the Centers for Disease Control and Prevention (CDC) (contract number 200-2020-09528), a grant from the Bill \& Melinda Gates Foundation, and a generous gift from The Flu Lab (https://theflulab.org).}
}

\maketitle

\begin{abstract}
Peripheral blood oxygen saturation (\spo2{}) is an essential indicator of respiratory functionality and received increasing attention during the COVID-19 pandemic. 
Clinical findings show that COVID-19 patients can have significantly low \spo2{} before any obvious symptoms. 
Measuring an individual's \spo2{} without having to come into contact with the person can lower the risk of cross contamination and blood circulation problems. 
The prevalence of smartphones has motivated researchers to investigate methods for monitoring \spo2{} using smartphone cameras. 
Most prior schemes involving smartphones are contact-based: They require using a fingertip to cover the phone's camera and the nearby light source to capture reemitted light from the illuminated tissue.
In this paper, we propose the first convolutional neural network based noncontact \spo2{} estimation scheme using smartphone cameras. The scheme analyzes the videos of an individual's hand for physiological sensing, which is convenient and comfortable for users and can protect their privacy and allow for keeping face masks on. We design explainable neural network architectures inspired by the optophysiological models for \spo2{} measurement and demonstrate the explainability by visualizing the weights for channel combination. 
Our proposed models outperform the state-of-the-art model that is designed for contact-based \spo2 measurement, showing the potential of the proposed method to contribute to public health. We also analyze the impact of skin type and the side of a hand on \spo2 estimation performance.
\end{abstract}


\footnote{2023 IEEE. Personal use is permitted, but republication/redistribution requires IEEE permission.}

\section{Introduction}

\IEEEPARstart{P}{eripheral} {blood oxygen saturation (\spo2{}) is an important physiological parameter that represents the level of {oxygenation} in the blood and reflects the adequacy of respiratory function~\cite{nitzan2014pulse}. The estimation and monitoring of \spo2{} are essential for the assessment of lung function and the treatment of chronic pulmonary diseases.

It has been reported that COVID-19 patients can present with clinically significant hypoxemia or drop in {blood} oxygen saturation, yet many do not exhibit respiratory symptoms~\cite{Couzin-Frankel455, starr2020pulse}. This highlights the importance of early detection of changes in \spo2 to facilitate timely management of asymptomatic patients with clinical deterioration.
{Conventional} \spo2{} measurement methods rely on contact-based sensing, including fingertip pulse oximetry and its variants in smartwatches and smartphones~\cite{severinghaus2007takuo, scully2011physiological, lu2015prototype, ding2018measuring}. These contact-based methods present the risk of {cross contamination} between individuals using the same measurement device. {An additional issue with contact-based methods is limb perfusion, especially in the digits. Circulation to the fingers and toes is often impaired in many  cardiovascular and pulmonary diseases, which complicates measurement of \spo2{}.} Also, pulse oximeters may not be widely available in {marginalized communities} and {some undeveloped} countries{~\cite{Herbert90}}.}

\rev{To provide a more comfortable and unobtrusive way to monitor \spo2{} that could be adopted in health screening and telehealth, a growing number of studies have investigated \spo2{} measurement using videos~\cite{kong2013non, van2016new, shao2015noncontact, tsai2016no, van2019data, bal2015non, casalino2020mhealth, tarassenko2014non, sun2021robust}, which allows for \spo2{} estimation without contact. However, \spo2 measurement using cameras in a contactless way, especially from smartphones, is challenging, because 1) the physiological signal is much weaker when acquired in a contactless setting compared the contact-based setting, and 2) the broadband absorption nature of the smartphone camera sensors lowers the optical selectivity to derive \spo2. Most prior art resorts to explicit feature extraction as a variant of the principle of pulse oximeters. Contrarily, in this paper, we introduce explainable convolutional neural network (CNN) models to extract features from all the three color channels holistically for \spo2 measurement from contactless videos captured using consumer-grade smartphone cameras. To the best of our knowledge, there is no prior work that remotely monitors \spo2{} with regular RGB cameras using neural networks.}

\begin{figure}[!t]

\hspace{-2mm}\includegraphics[width=1.02\linewidth]{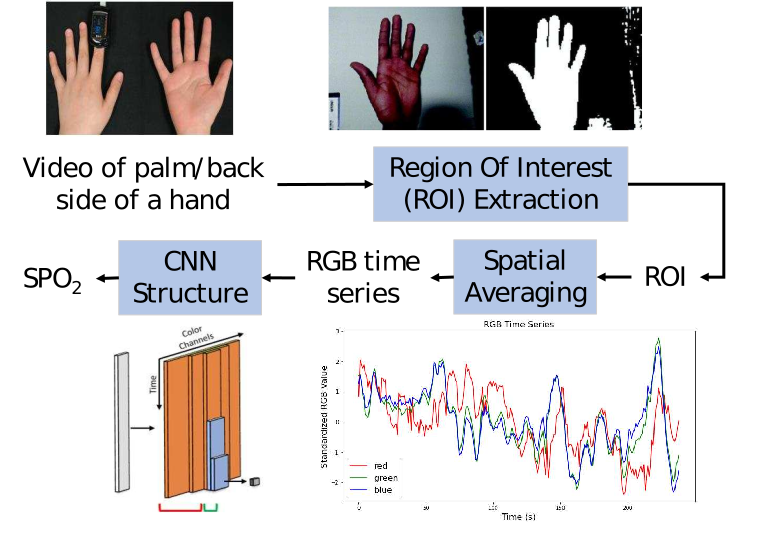}
\caption{
{\rev{The overall pipeline for our proposed method. The hand is first recorded and then the hand and skin pixels are segmented from the background. The average pixel values of the hand for each color channel is calculated frame by frame. This results in an RGB time series which is fed into the neural neural network model for \spo2{} estimation.}}
}
\label{fig:data processing}
\end{figure}

\rev{The overview of our proposed system for contactless \spo2{} monitoring using CNN from smartphones videos is shown in Fig.~\ref{fig:data processing}.} First, the region of interest (ROI), including the palm and {the} back of the hand, is extracted from the smartphone captured videos. Second, the ROI is spatially averaged to produce R, G, and B time series. Next, the three time series are fed into an optophysiology-inspired CNN for \spo2{} estimation\rev{, which is designed based on the light--tissue interaction principle 
applied to physiological sensing
and is aimed for better explainability of our proposed CNN.} We consider the hand region in this work as a proof-of-concept. Compared to using the face for \spo2 measurement as most of the prior art did~\cite{bal2015non, tarassenko2014non}, recording hand videos raises less privacy concern and is a safer way for {health condition screening and} data collection during the COVID-19 pandemic according to the mask-wearing guidelines. 

The contributions of our work are summarized as follows:
\begin{itemize}
\item To the best of our knowledge, this is the first work to use the synergy of all color channels and optophysiology-inspired \textbf{neural networks} to address the challenging problem of \textbf{contactless} \spo2{} sensing using consumer-grade RGB cameras.
\item Through a data-driven approach and visualization of the weights for the RGB channel combinations, we demonstrate the explainability of our model and that the choice of the color band learned by the neural network is consistent with the suggested color bands used in the optophysiological methods.
\item We analyze the impact of the two sides of the hand and different skin tones on the quality of \spo2{} estimation.
\item We achieve more accurate \spo2 prediction with our optophysiologically inspired neural network structures when compared to the state-of-the-art neural network structure designed for this problem.
\end{itemize}


\section{Background and Related Work}

\subsection{Blood Oxygen Saturation and the Ratio-of-Ratios Principle} 

\begin{figure}[!t]
\centering
\includegraphics[width=3.5in]{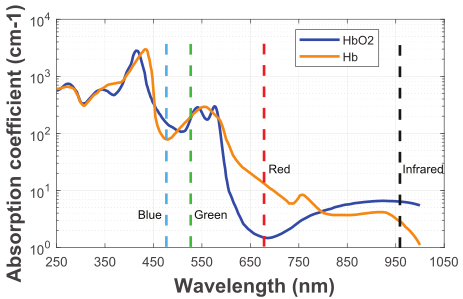}
\caption{Extinction coefficient curves of hemoglobin showing the absorption properties at different wavelengths. The curves were plotted based on \cite{ding2018measuring, hbcurve}. The difference between oxygenated hemoglobin (HbO$_2$) and deoxygenated hemoglobin (Hb) at the red and blue/infrared wavelengths means that these color channels contain 
useful information for \spo2{} prediction by means of optophysiological principles.}
\label{fig:extinction}
\end{figure}

The protein molecule hemoglobin (Hb) in the blood carries oxygen from the lungs to the tissues of the body. The level of blood oxygen saturation~(\spo2{}) represents the ratio of oxygenated hemoglobin~(HbO$_2$) to total hemoglobin and indicates the adequacy of respiratory function~\cite{nitzan2014pulse}. The normal range of \spo2{} is 95\% to 100\%~\cite{nitzan2014pulse}. \spo2{} is an important indicator of the ability of the respiratory system to meet metabolic demands. An abnormal drop in \spo2{} can serve as an early warning sign of inadequate oxygenation and clinical deterioration~\cite{nitzan2014pulse}. A convenient and noninvasive way to continuously measure \spo2{} is pulse oximetry~\cite{severinghaus2007takuo}. 

Pulse oximeters utilize the \textit{principle of the ratio of ratios} that was first proposed by Aoyagi in the early 1970s~\cite{severinghaus2007takuo}, and pulse oximeters are commonly used today in hospitals, clinics, and homes. The ratio-of-ratios method leverages the optical absorbance difference of Hb and HbO$_2$ at two wavelengths, which conventionally are at red and infrared wavelengths as indicated in Fig.~\ref{fig:extinction}. For the commonly seen pulse oximeters, lights at the two wavelengths are emitted through the fingertip. The transmitted light, interacted and attenuated by the blood and tissue, and received by an optical sensor, conveys information about pulsatile blood volume. The pulsatile blood volume at the two wavelengths is further processed to produce an \spo2{} estimation. \rev{Specifically, \spo2{} is proved to be approximately linear to the ratio-of-ratios, which is computed as the ratio between the quotients of pulsatile/AC component and relatively stationary/DC component of transmitted pulse signals at red and infrared wavelengths~\cite{severinghaus2007takuo}.}

\subsection{Video Based \spo2 Measurement}
\label{bkgd: video spo2}
With the prevalence of smartphones, researchers have investigated methods of monitoring \spo2{} using smartphones, most of which are contact-based and require the fingertips to be pushed against the illuminated light source and the built-in camera~\cite{scully2011physiological, lu2015prototype, ding2018measuring, nemcova, Lamonaca}, so that the diffusely reflected light by the fingertip is captured by the camera. In this setup, an adapted ratio-of-ratios model is utilized with the red and blue (or green) channels of color videos in lieu of the traditional narrowband red and infrared wavelengths. Those contact-based measurements can cause a sense of burning after several minutes of contact with the illuminated flashlight and is not suitable for sensitive skin or prolonged use.

\rev{Contactless \spo2{} measurement from videos, with weaker acquired pulse signal than contact sensing, are investigated to mitigate the issues from contact-based measurement. Based on the setup of cameras and light sources, existing noncontact, video-based \spo2{} estimation methods can be grouped into two main categories and both categories leverage the differences in the optical characteristics of Hb and HbO$_2$. Methods from the first category utilize monochromatic sensing similar to conventional pulse oximetry. They use either high-end monochromatic cameras with selected optical filters or controlled monochromatic light sources~\cite{kong2013non, van2016new, shao2015noncontact, tsai2016no}. The monochromatic light sources and sensors are selected so that they can have accurate control of the absorption effect of hemoglobins for precise \spo2{} measurement.
The other category uses consumer-grade RGB cameras that are more accessible than the monochrome setup, such as digital webcams and smartphone cameras~\cite{tarassenko2014non, bal2015non, casalino2020mhealth, sun2021robust}. \spo2{} measurement using those cameras are more challenging because compared to the narrowband photosensors in pulse oximeters, each R, G, and B channel of the regular photographic cameras used by smartphones today senses a much wider range of wavelengths, so the aggregation of the broadband signals lowers the difference of optical sensing between oxygenated vs. deoxygenated hemoglobins, making it much less optically selective compared to the narrowband sensing systems. To address the issue in this challenging scenario, most previous noncontact \spo2{} measurement sticked to the explicit feature extraction and linear regression as the variants of the two-color-channel based ratio-of-ratios methods~\cite{tarassenko2014non, bal2015non, casalino2020mhealth, sun2021robust}. In contrast, we address this issue by utilizing all the three color channels strategically. More specifically, in this paper, we use neural networks to distill the \spo2{} information from color channels in a holistic way.}

\subsection{Deep Learning Aided Camera-based Physiological Monitoring} {Deep learning has demonstrated promising performance in camera-based physiological {measurements}, such as heart rate, breathing rate, and body temperature~\cite{niu2019rhythmnet, chen2018deepphys, vspetlik2018visual, Zheng2020Mobile}. An end-to-end convolutional attention network was proposed in~\cite{chen2018deepphys} to estimate the blood volume pulse from face videos. Frequency analysis is then conducted on the estimated pulse signal for heart rate and breathing rate tracking. The study in~\cite{niu2019rhythmnet} demonstrates that the heart rate can be directly inferred using a convolutional network with spatial-temporal representations of the face videos as its input. Mobile applications have been developed {to} utilize CNNs to measure body temperature from facial images{~\cite{Zheng2020Mobile}}.}

Deep learning for \spo2{} monitoring from videos is still in {the} early stage. Ding~\textit{et al.}~\cite{ding2018measuring} {recently} proposed a convolutional neural network architecture for contact-based \spo2{} monitoring with smartphone cameras. Even though {the work in~\cite{ding2018measuring} showed} better performance than the conventional ratio-of-ratios method, their technique requires the users’ fingertips to be in contact with the illuminated flashlight and camera, which {not only may} lead to a sense of burning for a continuous period of time but also {raises} sanitation concerns, especially if the sensing device is shared by multiple participants during pandemics. Inspired by the optophysiological model for \spo2{} measurement~\cite{webster1997design, scully2011physiological, van2016new}, we develop a deep learning architecture to monitor \spo2{} in a contactless way with regular RGB cameras, which has the potential to be adopted in health screening and telehealth.

\rev{TABLE~\ref{method comp table} compares our proposed method with existing \spo2{} measurement methods. 
We detail the uniqueness of our method as follows.
Our contact-free method helps reduce the germ transfer and discomfort caused by the heat from contact with the flashlight on. 
It uses the hand as the ROI for better privacy protection. 
Its smartphone cameras based sensing is ubiquitous and convenient to use.
Its neural network's structure was inspired from light--tissue interaction principles, which makes it more explainable as compared to black box style neural networks. 
Lastly, the proposed method adopts an implicit way of feature extraction and model fitting instead of strictly following the conventional ratio-of-ratios principle, which can better cope with the more challenging scenario of \spo2{} sensing.}
 
\begin{table}[!t]
\centering
\rev{\caption{\rev{Comparison of various \spo2{} measurement methods.}}
\label{method comp table}
\resizebox{3.5in}{!}{%
\begin{tabular}{c|ccccc}
\hline\hline
\multirow{3}{*}{\begin{tabular}[c]{@{}c@{}}\textbf{Methods} \end{tabular}} & \multicolumn{5}{c}{\textbf{Characteristics}} \\ \cline{2-6}  &  \begin{tabular}[c]{@{}c@{}} \textbf{Non-}\\ \textbf{contact?}  \end{tabular} &
\begin{tabular}[c]{@{}c@{}} \textbf{ROI}\\  \end{tabular} & \begin{tabular}[c]{@{}c@{}} \textbf{Measurement}\\ \textbf{Equipment}  \end{tabular} & \begin{tabular}[c]{@{}c@{}} \textbf{Optophy-}\\ \textbf{siology?} \end{tabular} & \begin{tabular}[c]{@{}c@{}} \textbf{Feature}  \\ \textbf{Extraction} \end{tabular} \\ \hline
\cite{scully2011physiological, lu2015prototype} & \multirow{2}{*}{\xmark} & \multirow{2}{*}{Fingertip} & \multirow{2}{*}{Smartphone} & \multirow{2}{*}{\cmark} & \multirow{2}{*}{Explicit} \\
\cite{nemcova,Lamonaca}
\\\hline
\cite{ding2018measuring} & \xmark & Fingertip & Smartphone & \xmark & Implicit (CNN) \\\hline
\multirow{2}{*}{\cite{kong2013non, van2016new, shao2015noncontact}} & \multirow{2}{*}{\cmark} & \multirow{2}{*}{Face} & Mono. Cameras/ & \multirow{2}{*}{\cmark} & \multirow{2}{*}{Explicit} \\
&&&Light Source
\\\hline
\cite{tarassenko2014non, bal2015non, casalino2020mhealth, sun2021robust} & \cmark & Face & RGB cameras & \cmark & Explicit  \\\hline
\multirow{2}{*}{\cite{tsai2016no}} & \multirow{2}{*}{\cmark} & \multirow{2}{*}{Hand} & Mono. Cameras/ & \multirow{2}{*}{\cmark} & \multirow{2}{*}{Explicit} \\
&&& Light Source
\\\hline
\textit{Proposed} & \multirow{2}{*}{\cmark} & \multirow{2}{*}{Hand} & \multirow{2}{*}{Smartphone} & \multirow{2}{*}{\cmark} & \multirow{2}{*}{Implicit (CNN)} \\
\textit{Method}
\\ \hline\hline
\end{tabular}}}
\end{table}

\section{Proposed Method for \spo2{} From Videos}

We aim to estimate \spo2{} levels using a hand video {based on} the fact that the color of the skin changes subtly when red cells in the blood flow carry/release oxygen.
Our proposed method, as illustrated at a high level in Fig.~\ref{fig:data processing}, extracts three color time-series from the skin area of the hand video.
The extracted time series are then fed to optophysiology-inspired neural networks designed to achieve more explainable {and accurate} \spo2{} predictions.

\subsection{Extraction of Skin Color Signals}

The physiological information related to \spo2{} is embedded in the color of the reflected/reemitted light from a person's skin.  
Hence, a preprocessing step that precisely extracts the color information from the skin area is crucial to the design of an effective \spo2 estimation method.
For each participant's video, we aim to extract the R, G, and B time series and refer to these 1-D time series as \emph{skin color signals}.
We first need to locate the ROI of the skin pixels from the video. 
We {have} found that it is most effective to discriminate the skin pixels from the background along the Cr {axis} of the YCbCr color space \cite{Burger2008Digital}.

We use Otsu's method~\cite{otsu1979threshold} to determine a threshold that best separates the skin pixels from the background by minimizing the variance within the skin and non-skin classes. Once the ROI corresponding to the hand is located, the R, G, and B time series are generated by spatially averaging over the values of skin pixels for each frame of the video.

The skin color signals are split up into 10-second segments using a sliding window with a step size/stride of 0.2 seconds to serve as the inputs for neural networks. From an optophysiological perspective, the reflected/reemitted light from the skin 
for the duration of one cycle of heartbeat, i.e., 0.5--1 seconds for a heart rate of 60--120 bpm,
should contain almost the complete information necessary to estimate the instantaneous \spo2{}~\cite{severinghaus2007takuo}.
In our system design, we use longer segments to add resilience against sensing noise.
Since the segment length is one order of magnitude longer than the minimally required length to contain the \spo2{} information, we can use a fully-connected or convolutional structure to adequately capture the temporal dependencies without resorting to a recurrent neural network structure.

\subsection{Neural Network Architectures}

\begin{figure*}[!t]
\centering
\includegraphics[width=7in]{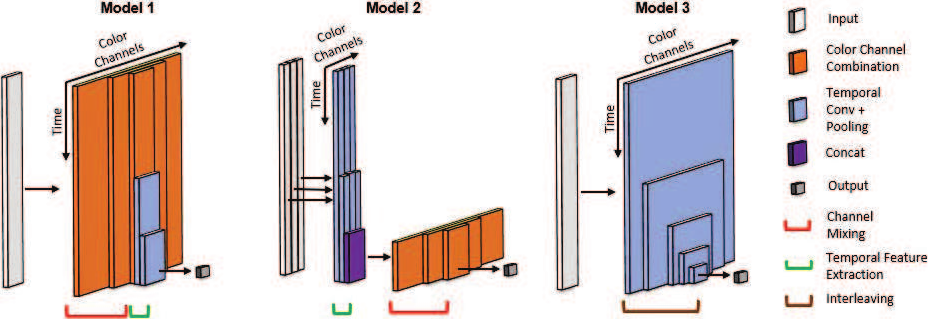}
\caption{
Proposed network structures for predicting an \spo2{} level from a fixed-length segment of skin color signals. We highlight the differences among {the} three model configurations instead of showing the exact model structures.
Model~1 combines the RGB channels before temporal feature extraction. Model~2 extracts the temporal features from each channel separately and fuses them toward the end. Model~3 interleaves color channel mixing and temporal feature extraction
.}
\label{fig:structures}
\end{figure*}

The previous neural network work for \spo2{} prediction mainly explored prediction, but not the model explainability~\cite{ding2018measuring}. Explainability/interpretability is highly desirable in many applications yet often not sufficiently addressed, partly due to the black box nature of neural networks.
From a healthcare standpoint, explainability is a key factor {that} should be taken into account at the beginning of the design of a system. \rev{Explainability is important for physiological and healthcare based applications to validate and provide strong justifications to users/clinicians that the proposed approach has a sound rationale and is trustworthy. We have learned this firsthand from discussions with collaborating medical professionals. Confirming that the neural network model is learning clinically relevant information will help ensure that the estimation capability can be replicated by others. This can be concerning if the performance varies too significantly across different subjects. Explainability may also be necessary for legal, certification, and regulatory purposes. Explainability in publicly-funded research is also widely sought by governmental funding agencies to cope with taxpayers' increasing expectations of transparency.} 

To extract features from the skin color signals and estimate \spo2{}, we propose three physiologically motivated neural network structures. These structures are inspired by 
domain knowledge-driven
physiological sensing methods and designed to be physically explainable. 
For heart rate sensing~\cite{zhu2017fitness, niu2019rhythmnet} and respiratory rate sensing~\cite{9154523},
the RGB skin color signals are often combined first followed by temporal feature extraction, as is done in the plane-orthogonal-to-skin (POS) algorithm~\cite{POS}. In contrast, for conventional \spo2{} sensing methods such as the ratio-of-ratios~\cite{webster1997design}, the temporal features are extracted first and the color components are combined at the end. Our proposed neural network structures explore different arrangements of channel combination and temporal feature extraction. We want to systematically compare the performance of our explainable model structures.

\vspace{1mm}\noindent{}\textbf{{Color} Channel Mixing Followed by Feature Extraction.} In Model~1, shown as the leftmost structure depicted in Fig.~\ref{fig:structures}, we combine the color channels first using several channel combination layers and then extract temporal features using temporal convolution and max pooling. A channel combination layer first linearly combines the $C_\text{in}$ input channels/vectors into $C_\text{out}$ activation vectors and then applies a rectified linear unit (ReLU) activation function to obtain the output channels/vectors. Mathematically, the channel combination layer is described as follows:
\begin{equation}
    \mathbf{V}  =  \sigma(\mathbf{W} \mathbf{U}  + \mathbf{b \mathbbm{1}}^T),
\end{equation}
\noindent{}where $\mathbf{U} \in \mathbb{R}^{C_\text{in} \times L}$ is the input comprised of $C_\text{in}$ time series/vectors of length $L$. The initial channel combination layer has an input of three channels with 300 points along the time axis. $\mathbf{W} \in \mathbb{R}^{C_\text{out} \times C_\text{in}} $ is a weight matrix, where each of the $C_\text{out}$ rows of the matrix is a different linear combination for the input channels. 
A bias vector $\mathbf{b} \in \mathbb{R}^{C_\text{out}}$ contains the bias terms for each of the $C_\text{out}$ output channels, which ensures that each data {point} in the created segment of length $L$ has the same intercept. $\mathbbm{1}^T\in \R^{1 \times L}$ is a row vector of all ones.
The nonlinear ReLU function $\mathbb{\sigma}(x) = \max(0, x)$ is applied elementwise to the activation map/matrix. The output of the channel combination layer $\mathbf{V} \in \mathbb{R}^{C_\text{out} \times L}$ contains $C_\text{out}$ channels of nonlinearly combined input channels. 

The channel mixing section concatenates multiple channel combination layers with decreasing channel counts to provide significant nonlinearity. The output of the last channel combination layer has seven channels. After the channel mixing, for temporal feature extraction, we utilize multiple convolutional and max pooling layers with a downsampling factor of two to extract the temporal features of the channel-mixed signals. When there are multiple filters in the convolutional layer, there will also be some additional channel combining with each filter outputting a channel-mixed signal. Finally, a single node is used to represent the predicted \spo2{} level. This model has three channel combination layers, three feature extraction layers, and {a total of} 34K trainable parameters.

\vspace{1mm}\noindent{}\textbf{Feature Extraction Followed by {Color} Channel Mixing.} In Model~2, {which is} the middle structure depicted in Fig.~\ref{fig:structures}, we reverse the order of {color} channel mixing and temporal feature extraction from that in Model 1. The three color channels are separately fed for temporal feature extraction. The convolutional layers learn different features unique to each channel. At the output of the temporal feature extraction section, each color channel has been downsampled to 
retain only the important temporal information.
The color channels are then mixed together in the same way as described 
for Model~1 before outputting the \spo2{} value. This model has three channel combination layers, 2 feature extraction layers, and {a total of} 12K parameters.

\vspace{1mm}\noindent{}\textbf{Interleaving Feature Extraction and Channel Mixing.} In our third model, we explore the possibility of interleaving the color channel mixing and temporal feature extraction stpng. As illustrated by the rightmost structure depicted in Fig.~\ref{fig:structures}, the input is first put through a convolutional layer with many filters and then passed to max pooling layers, resulting in feature extraction along the time as well channel combinations through each filter. The number of filters is reduced with each successive convolutional layer, gradually decreasing the number of combined channels and downsampling the signal in the time domain. This model has 4 layers and {a total of} 307K parameters. 

\vspace{1mm}\noindent{}\textbf{Loss Function and Parameter Tuning.} We use the  root-mean-squared-error (RMSE) as the loss function for all models. During training, we save the model instance at the epoch {that has} the lowest validation loss. The neural network inputs are scaled to have zero mean and unit variance to improve the numerical stability of the learning. The parameters and hyperparameters of each model structure were tuned using the HyperBand algorithm~\cite{li2018hyperband}, which allows for faster and more efficient search over a large parameter space than grid search or random search. It does this by running random {parameter} configurations on a specific schedule of iterations per configuration, and uses earlier results to select candidates for longer runs. The parameters that {are} tuned 
include the learning rate, the number of filters and kernel size for convolutional layers, the number of nodes, the dropout probability, and whether to do batch normalization after each convolutional layer.

\section{Experimental Results} 
\label{experiment result}
\subsection{Dataset and Capturing Conditions}

\begin{figure}[!t]
\centering
\includegraphics[width=3in]{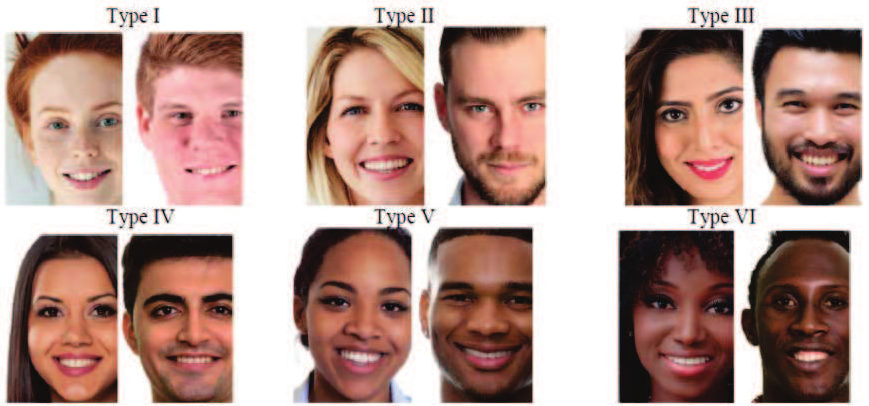}
\caption{{Fitzpatrick skin types~\cite{skin-types}.}}
\label{fig:skin types}
\end{figure}

Our proposed models were evaluated on a self-collected dataset. The dataset consisted of hand video recordings and \spo2{} data from fourteen participants, of which there were six males and eight females between the ages of 21 and 30. Participants were asked to categorize their skin tone based on the Fitzpatrick skin types \cite{skin-types} {shown in Fig.~\ref{fig:skin types}}.  
The distribution of the participants’ skin types is as follows:
Two participants of type II, eight participants of type III, one participant of type IV, and three participants of type V. This research was using protocol {\#1376735} approved by the University of Maryland Institutional Review Board (IRB).

\begin{figure}[!t]
\centering
\includegraphics[width=2in]{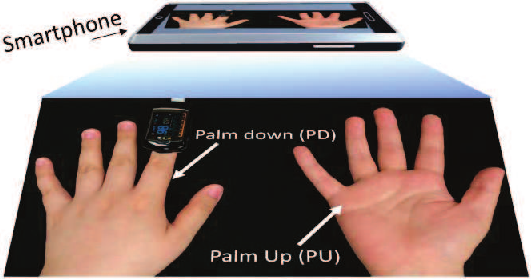}
\caption{
\rev{Setup for capturing hand videos. Over the top is an smartphone device recording a single video for both hands.
The left and right hands are in the \textit{palm down~(PD)} and \textit{palm up~(PU)} positions, respectively. The CMS-50E pulse oximeter clamped to the left index finger records reference \spo2{} signals.} 
}
\label{fig:setup}
\end{figure}

\begin{figure}[!t]
  \centering

  \subfloat[]{\hspace{-1mm}\includegraphics[width=1.7in, height=1.05in]{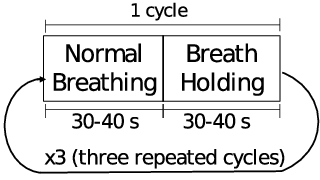}\label{fig:breathing}}

  \subfloat[]{\includegraphics[width=1.7in]{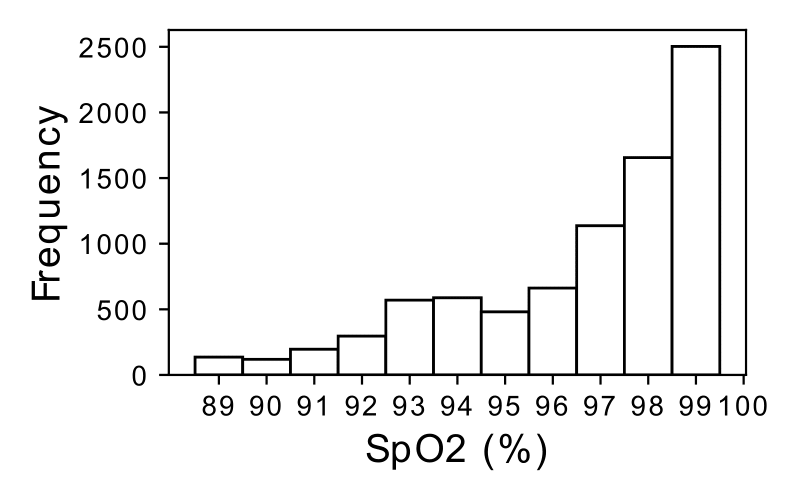}\label{fig:spo2 distr}}
  \hfill
  \caption{(a) Breathing protocol that participants were asked to follow, including 3 cycles of normal breathing and breath holding. (b) Histogram of \spo2{} values in the collected dataset.
  }
 
\end{figure}

During data collection, each participant was asked to place his/her hands still on a table to avoid hand motion. Their palm of the right hand and the back of the left hand were facing the camera, as illustrated in Fig.~\ref{fig:setup}. We refer to these two hand-video capturing positions as \emph{palm up~(PU)} and \emph{palm down~(PD)}, respectively. Each participant was asked to follow the breathing protocol outlined in Fig.~\subref*{fig:breathing}. \rev{In each \textit{breath-holding cycle}, the participant breathes normally for 30--40 seconds, exhales all the way, and then holds his/her breath for another 30--40 seconds. This breath-holding protocol aims to induce a wide dynamic range of \spo2{} levels. The normal \spo2{} range for a healthy person is between $95\%~\text{and}~100\%$. By holding breath, \spo2{} can drop below $90\%$, as shown in the distribution of \spo2{} values in Fig.~\subref*{fig:spo2 distr}.} \rev{This breath-holding cycle is repeated three times in each recording and each participant has two recordings with at least 15 minutes in between for both PU and PD hand-video capturing positions, resulting in a total of 56 recordings considering PU and PD separately.} All videos were recorded using an iPhone 7 Plus. The participant’s \spo2{} was simultaneously measured using a CONTEC CMS-50E pulse oximeter clamped to the left index finger of the hand. We use this pulse oximeter as the reference measurement as it has been validated to be within $\pm2\%$ of the true \spo2{} level for the range of \spo2{} levels in our dataset. \rev{The video frame rate of the smartphone is 30~fps and the sampling rate of the pulse oximeter for the reference \spo2{} measurements is 1~Hz.}

The reference \spo2{} signal is interpolated to 5 sample points per second to match the segment sampling rate using a smooth spline approximation \cite{spline}. Each RGB segment and \spo2{} value pair is fed into our models as a single data point, the models output a single \spo2{} estimate per segment. To evaluate a model {on a video recording}, the model is sequentially fed {with} all RGB segments from the recording to generate a time series of preliminarily predicted \spo2{} values. All predictions greater than 100\% \spo2{} are clipped to 100\% {based on physiological knowledge}. A 10-second long moving average filter is applied to generate a refined time series of predicted \spo2{} values.

\subsection{Participant-Specific Results}
\label{sec: PS}
To investigate how well 
the proposed models could learn to estimate a specific individual's \spo2{} from his/her own data, we first conducted participant-specific experiments, that is, we learn individualized models for each participant. \rev{The participant-specific experiment is important because 1)~it sets a baseline understanding of how a model performs when the participant is known, where we can develop a specific model for him/her to improve personalized healthcare; 2)~it enables the analysis of the health status of an individual over time and contributes to the emerging \textit{digital twins} paradigm for personalized healthcare~\cite{bruynseels2018digital}.}

\begin{figure}[!t]
  
  \centering
  \subfloat[]{\includegraphics[width=\linewidth]
  {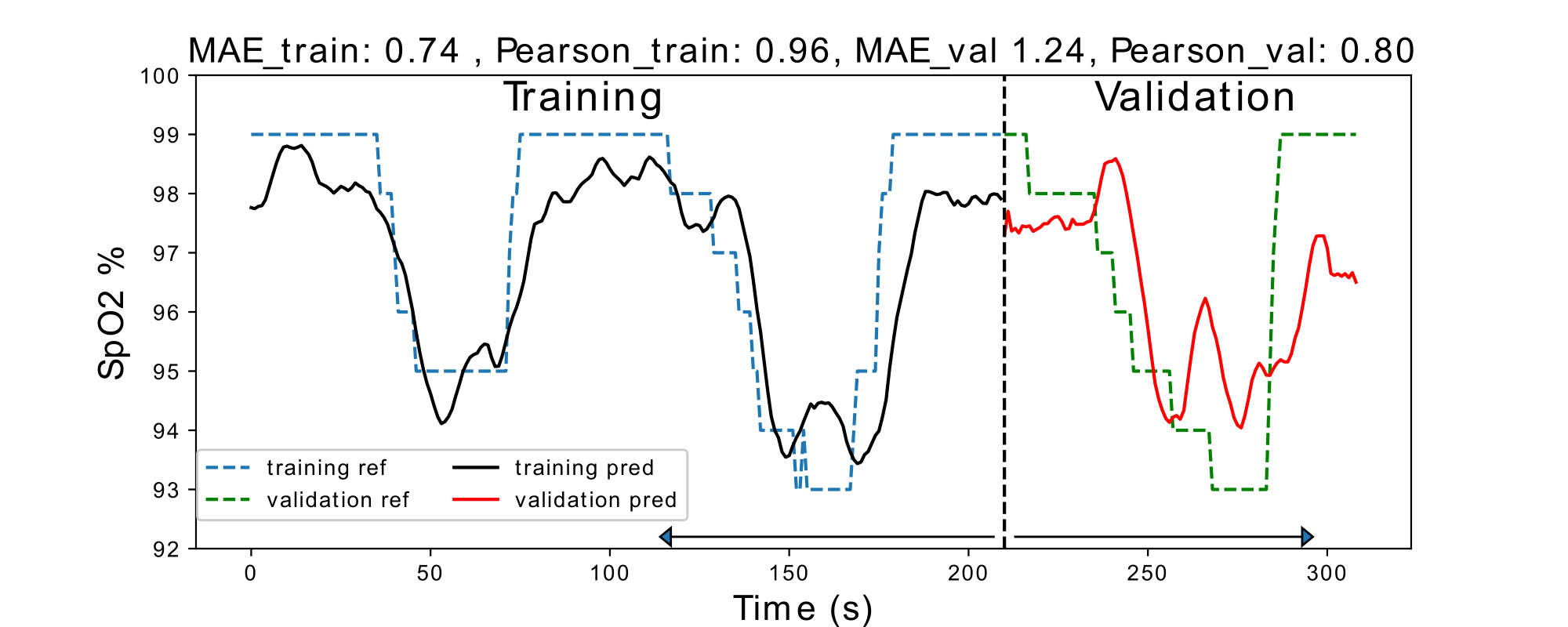} \label{fig:train-val}} \\

  \subfloat[]{\includegraphics[width=\linewidth]
  {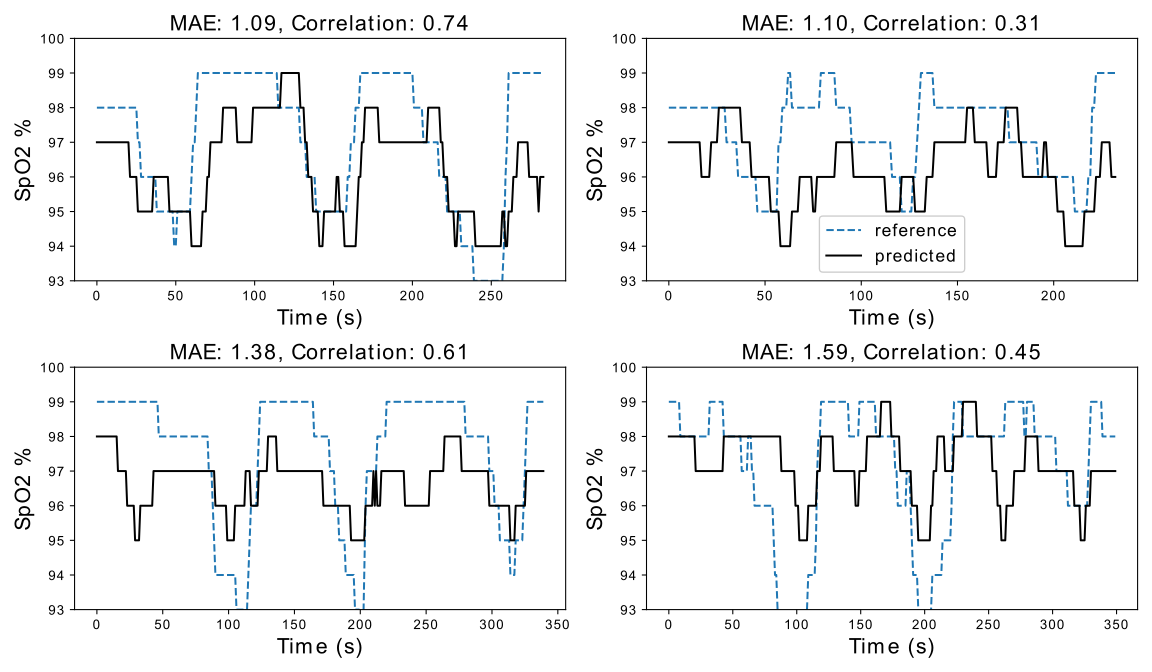} \label{fig:plots}}
  
  \caption{(a) Training vs. validation predictions. (b) Test predictions of varying performance with reference \spo2. The higher the Pearson's correlation, the better the {prediction} captures the reference \spo2 trend. The lower the MAE, the better the {prediction captures} the dips in \spo2.
  }
  
  \label{fig:AB}
\end{figure}

\begin{figure}[!t]
  \centering
  \subfloat[]{\hspace{-2mm}\includegraphics[width=0.5\linewidth]{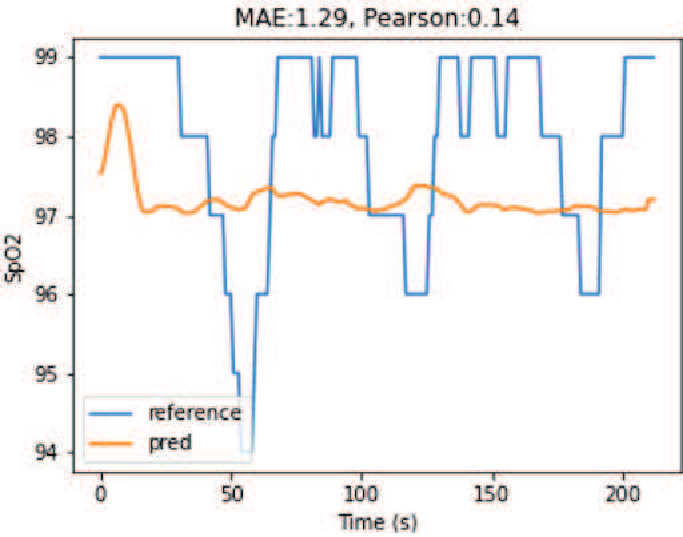}\label{fig:bad_1}}
  
  \subfloat[]{\includegraphics[width=0.5\linewidth]{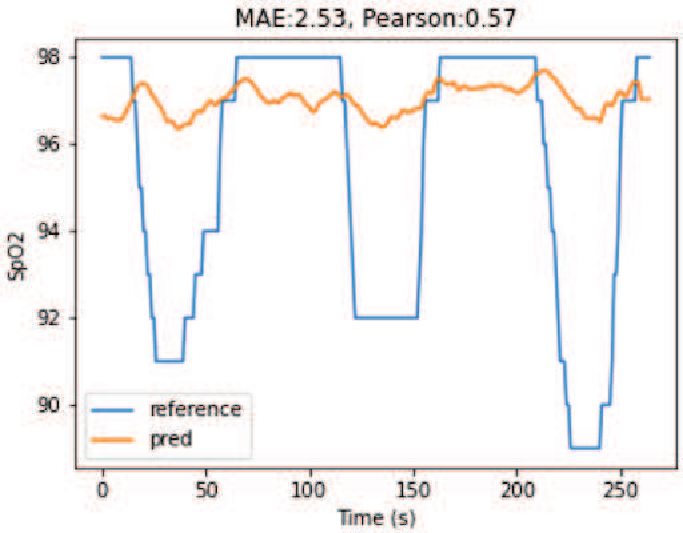}\label{fig:bad_2}}
 
  \caption{\rev{Cases where the proposed model fails to accurately predict \spo2{}. (a) The model predicts a near constant \spo2{} that results in small MAE but also low correlation. (b) The model prediction follows the general trend of the reference \spo2{} signal but does not capture the magnitude of the dips. This results in a high correlation but also large MAE.
  }}
\end{figure}

\noindent \textbf{Experimental Setting.}
\rev{Since each participant has two recordings for each of the PU and PD cases, one recording is used for training and validation of the model and the 
other
recording is for testing.} An example of the training and validation predictions curves is shown in Fig.~\subref*{fig:train-val}. Each recording contains three breathing cycles, for each training/validation recording, the first two breathing cycles are taken for training and the third cycle is used for validation. Splitting the recordings into cycles instead of randomly sampling the 10-sec overlapping RGB segments ensures that there are no overlapping segments of data between the training and validation set. Example test prediction curves and their \rev{Pearson's correlation and mean absolute error~(MAE) are shown for reference in Fig.~\subref*{fig:plots}. The Pearson's correlation shows how well the predicted signal follows the rising and falling trend in the reference signal.} 
\rev{It should be noted that if the correlation is low, e.g., a constant temporal estimate, then the MAE and RMSE metrics are less meaningful as shown in Fig.~\subref*{fig:bad_1}. It is also possible for the correlation to be high but the absolute difference between the predicted and reference signal is large as shown in Fig.~\subref*{fig:bad_2}}. For the participant-specific experiments, due to the small dataset size, we augment the training and validation data by sampling with replacement. {This is an example of the bootstrapping data reuse strategy ~\cite[Chapter~5]{jamesICLR}.} The oversampling also helps address the imbalance in \spo2 data values that is shown in Fig.~\subref*{fig:spo2 distr}.

In each experiment, the model structure and hyperparameters are first tuned using the training and validation data. Once the model has been tuned, we train multiple instances of the model using the best tuned hyperparameters. Between each instance, we vary the random seed used for model weights initialization and random oversampling. Each model instance is evaluated on the training/validation recording, the model instance that achieves the highest validation RMSE is selected for evaluation on the test recording. This model is then evaluated on the test recording to obtain the final test results.

\noindent\textbf{Results.}
\rev{TABLE~\ref{tab:SS} shows the performance comparison of our proposed models with the prior-art model from Ding~\emph{et al.}~\cite{ding2018measuring} and the classic ratio-of-ratios method proposed by Scully~\emph{et al.}~\cite{scully2011physiological}. To the best of our knowledge, Ding~\emph{et al.}'s model is the only convolutional neural network structure that has been proposed for contact-based \spo2{} estimation. Its structure is similar to our Model~3 but with fewer layers. We select~\cite{scully2011physiological} to compare because other noncontact \spo2{} methods listed in TABLE~\ref{method comp table}, such as~\cite{kong2013non, shao2015noncontact,tsai2016no,tarassenko2014non, bal2015non, casalino2020mhealth, sun2021robust}, are variants of it with different light source/sensor setups while their algorithmic core is still based on the ratio-of-ratios principles that requires explicit AC/DC feature extraction and machine learning algorithms to model the relation between the values of the ratio-of-ratios and \spo2{}.} 

The performance is measured in Pearson's Correlation, mean absolute error~(MAE), and root mean squared error~(RMSE) and results of each condition are summarized in the median and interquartile range~(IQR). IQR quantifies the spread of an empirical distribution of a set of data points by computing the difference between the first quartile and the third quartile of the distribution. \rev{We use IQR to describe the middle spread of the data as well as to distinguish outliers.}

\begin{table}[!t]
\caption{Performance comparison of each model structure for participant-specific experiments. Results are given as the test median and IQR of all participants.}
\resizebox{3.4in}{!}{%
\begin{tabular}{cc|cc|cc|cc}
\hline
\multicolumn{1}{l}{} & \multicolumn{1}{l|}{Hand} & \multicolumn{2}{c|}{Correlation} & \multicolumn{2}{c|}{MAE (\%)} & \multicolumn{2}{c}{RMSE (\%)} \\
 & Mode & Median & IQR & Median & IQR & Median & IQR \\ \hline \hline
Model 1 & PD &  0.41 & 0.40 & {2.12}& 0.91 &{2.51}& 0.78\\
 (Proposed) & PU &  0.39 & 0.37 & 2.16 & 1.80 & 2.70 & 2.09\\ \hline
Model 2 & PD &  \textbf{0.46} & 0.44  & 2.09 & 1.32 & 2.52& 1.63 \\
(Proposed) & PU & \textbf{0.41} & 0.32& 1.96 & 0.68 & 2.48 & 0.89 \\ \hline
Model 3 & PD & {0.44} & 0.40  & \textbf{1.93} & 1.11 & {2.48} & 1.31 \\
(Proposed) & PU &  \textbf{0.41} & 0.46 & \textbf{1.81} & 1.83 & {2.43} & 2.44 \\ \hline \hline
\multirow{2}{*}{Scully \emph{et al.} \cite{scully2011physiological}} & PD & 0.08 & 0.37 & 1.94  & 0.92 & \textbf{2.22}  & 0.77 \\
 & PU & 0.19 & 0.24 & 2.01 & 0.80  & \textbf{2.36} & 0.78 \\ \hline
\multirow{2}{*}{Ding \emph{et al.} \cite{ding2018measuring}} & PD &  0.38 & 0.39& 3.25 & 2.85& 3.83 &3.24 \\
 & PU &  0.34 & 0.56 & {3.40} &3.16 & 4.58 &3.12\\
 \hline
\end{tabular}%

}
\label{tab:SS}
\end{table}

TABLE~\ref{tab:SS} reveals that Model~2 {achieves the best correlation} in both PD and PU cases, whereas Model~3 achieves the best MAE and a comparable correlation with Model~2, suggesting that Model~2 and Model~3 are comparably the best in the individualized learning. {Even though the method proposed in Scully~\emph{et al.}~\cite{scully2011physiological} achieves the {best (lowest)} RMSE, its correlations are {the worst (lowest)}. This suggests that the classic ratio-of-ratios method cannot track the trend of \spo2 well {using} the contactless {measurement by smartphone.}} All of our model configurations outperform Ding~\emph{et al.}~\cite{ding2018measuring}. For example, in the PU case for Model~3, the correlation is improved from 0.34 to 0.41 and the MAE is lowered from 3.40\% to 1.81\%. It is worth noting that the international standard for clinically acceptable pulse oximeters {allows} an error of 4\%~\cite{ISO}, and our estimation errors are all within this range.

\begin{figure}[!t]
  \centering
  \subfloat[]{\hspace{-2mm}\includegraphics[width=0.59\linewidth]{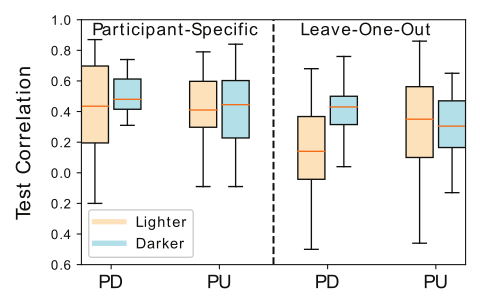}\label{fig:skin}}
  
  \subfloat[]{\includegraphics[height=0.38\linewidth]{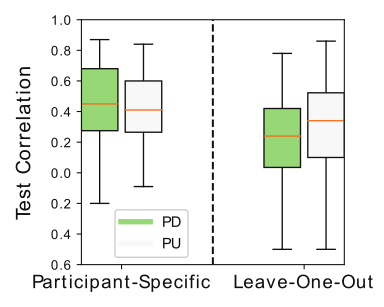}\label{fig:pd_pu}}
 
  \caption{Box plots comparing distributions of correlations for (a) lighter vs. darker skin types, and (b) PD vs. PU for all skin types. The PD results are better for darker skin tones in both the participant-specific and leave-one-out cases. 
  }
  \label{fig:boxes}
\end{figure}

{There are two factors,} including the skin type and the side of the hand, {which} might influence the performance of \spo2{} estimation. {We therefore investigate} the following two questions: (1) Whether the different skin types matter in PU or PD cases, and (2) whether the side of hand matters in lighter skin (types II + III) or darker skin (types IV + V). The box plots in Fig.~\ref{fig:boxes} {summarize} the distributions of the test correlations from all the three proposed models in PU and PD modes of (a) lighter-skin and darker-skin participants, and (b) all participants.

\noindent{\textbf{Bayesian statistical test:} We use Bayesian statistical tests to further analyze the results in Fig.~\ref{fig:boxes} by providing a probabilistic assessment of whether the results from two groups being compared have the same mean~\cite{kruschke2018bayesian2, kruschke2013bayesian,kruschke2018bayesian1,kruschke2021bayesian,kruschke2018rejecting}. We avoid using the popular $t$-test because it makes only a binary decision due to its lack of direct information about the probability of difference between group means {of the given data}~\cite{kruschke2013bayesian}. In contrast, the Bayesian statistical test {computes} the posterior distribution of difference {between the} two group means {to} quantify its certainty of possible values~\cite{kruschke2018bayesian2}. The decision rule of the Bayesian statistical test for the null hypothesis that the two groups have the same mean can be stated as follows given the \textit{region of practical equivalence (ROPE)} of zero difference~\cite{kruschke2021bayesian}:
\begin{itemize}
    \item (\textit{Accepted}): If the percentage of the posterior distribution of the group-mean difference inside the ROPE is sufficiently high (e.g., greater than 95\%~\cite{kruschke2021bayesian}), then the null hypothesis is accepted.
    \item (\textit{Rejected}): If the percentage of the posterior distribution of the group-mean difference inside the ROPE is sufficiently low (e.g., less than 2.5\%~\cite{kruschke2021bayesian}), then the null hypothesis is rejected.
    \item (\textit{Undecided}): When the null hypothesis is neither accepted nor rejected, the percentage of the posterior distribution of the group-mean difference inside the ROPE can be used to quantify the certainty that two group means are the same. One example is shown in Fig.~\ref{fig:bayesian_example}.
\end{itemize}
\begin{figure}[!t]
  \centering

  \includegraphics[width=3.5in]{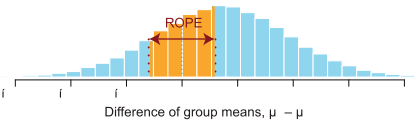}
  \caption{{Posterior distribution of the difference of group means. This shows an example of an undecided case of the Bayesian statistical test given that the ROPE of zero difference is {set to $[-0.03,0.03]$ and 33\%} of the posterior distribution falls within the ROPE. {The percentage of coverage} can be used to quantify the certainty that two groups have the same mean.}}
  \label{fig:bayesian_example}
\end{figure}
To conduct the Bayesian statistical tests, we use an R statistical package {named} BEST~\cite{R-manual-for-Best}. 
To determine the ROPE on the difference between the means, we use Cohen's established convention that the ROPE of small standardized mean difference is $[-0.1,0.1]$~\cite{kruschke2018bayesian1,kruschke2018rejecting}. {Given the standard deviation $0.3$ of our data, the ROPE for the difference of means of our data is scaled to $[-0.03, 0.03]$.}}

{To answer question (1) about the impact of the skin type on the prediction performance, we focus on the left panel of Fig.~\subref*{fig:skin}.
For the PD case, {only 14\% of the posterior distribution of the difference between the means of the lighter and darker skin groups falls in the ROPE.} For the PU case, {23\% of the posterior distribution {falls in} the ROPE. This suggests that it is highly credible to conclude that the skin type makes a difference for \spo2 prediction, and the difference is more certain {to be observed} when using the {back} of the hand as the ROI compared to using the palm.}

To answer question~(2), we first focus on the left panel of Fig.~\subref*{fig:pd_pu} {when participants of all skin colors are considered together. 33\% of the posterior distribution of the difference of means between PU and PD cases falls in the ROPE.} {We then} zoom into the darker skin group as shown in the left panel of Fig.~\subref*{fig:skin}, {only 15\% of the posterior distribution of the difference of means between PD and PU cases falls in the ROPE, whereas for the lighter skin group, 31\% of the posterior distribution {falls} in the ROPE. This implies that it is highly credible that the side of the hand may have some impact on \spo2 prediction, especially when concerning mainly the darker skin group.}}

\subsection{Leave-One-Participant-Out Results}

To investigate whether the features learned by the model from other participants are generalizable to new participants whom it has not seen before, we conduct leave-one-participant-out experiments. For each experiment, when testing on a certain participant, we use all the other participant’s data for training and leave the test participant’s data out. The recordings from all the non-test participants are used for participant-wise cross-validation to select the best model structure and hyperparameters. The selected model is evaluated on the two recordings of the test participant, whose data was never seen by the model during training. 

TABLE~\ref{tab:loo table} shows the performance comparison of each model in leave-one-participant-out experiments. Model~1 achieved the best performance in terms of correlation and achieved the best MAE and RMSE for the PU case. {Similar to the participant-specific case, the classic ratio-of-ratios method proposed in Scully~\emph{et al.}~\cite{scully2011physiological} achieved better MAE and RMSE results for the PD case but the correlation result was low, suggesting that the model achieved low error by simply predicting a nearly constant \spo2{} near the middle of the \spo2{} range.} The best performance of Model~1 in the leave-one-participant-out experiment may imply that the features extracted after combining the color channels at the beginning of the pipeline can be generalized better to unseen participants than the features extracted before channel combination or through interleaving as in Models~2~or~3.

In the participant-specific case, the model is specifically tailored to the test individual, whereas the leave-one-participant-out case is more difficult because the model needs to accommodate for the variation in the population. As expected, in Fig.~\ref{fig:boxes}, we observe that the overall results from the leave-one-participant-out experiments do not match those from the participant-specific experiments. Because of the modest size of the dataset, the model has not seen as diverse data as a larger and richer dataset would offer. The generalization capability to new participants can be improved when more data is available.

\begin{table}[!t]
\caption{Performance comparison of each model structure in leave-one-participant-out experiments. Results are given as the test median and IQR of all participants.
}
\centering
\resizebox{3.4in}{!}{%
\begin{tabular}{cc|cc|cc|cc}
\hline
\multicolumn{1}{l}{} & \multicolumn{1}{l|}{Hand} & \multicolumn{2}{c|}{Correlation} & \multicolumn{2}{c|}{MAE (\%)} & \multicolumn{2}{c}{RMSE (\%)} \\
 & Mode & Median & IQR & Median & IQR & Median & IQR \\ \hline \hline
{Model 1} & PD & \textbf{0.33} &0.42 & {2.33} & 1.07& 3.07 &1.52 \\
(Proposed) & PU & \textbf{0.46} &0.36 & \textbf{1.97} & 0.80& \textbf{2.32} &0.87 \\ \hline
{Model 2} & PD & 0.15 &0.50 & 2.43 & 0.94& 3.35  & 1.11 \\
 (Proposed) & PU & 0.33 & 0.39& 2.08 & 0.73& 2.41 & 0.71\\ \hline
{Model 3} & PD & 0.23 & 0.38& {2.48} & 1.18& {2.98} &1.33 \\
(Proposed) & PU & 0.27 & 0.31& 2.02 & 1.03& 2.54 & 1.28\\ \hline \hline
\multirow{2}{*}{Scully \emph{et al.} \cite{scully2011physiological}} & PD & 0.05& 0.43& \textbf{2.08}  & 0.65& \textbf{2.44}  & 1.14\\
 & PU & 0.01 &0.54 &2.08 & 0.60 & 2.43 &1.20 \\ \hline
\multirow{2}{*}{Ding \emph{et al.} \cite{ding2018measuring}} & PD & 0.11 &0.56 & 3.19 & 1.61& 3.76 & 1.52\\
 & PU & 0.26 &0.42 & 2.43 & 1.22 & 2.85 &1.51 \\ \hline
\end{tabular}%
}
\label{tab:loo table}
\end{table}

{{We now revisit} the two research questions raised in Section~\ref{sec: PS} under the leave-one-participant-out scenario. 
First, we analyze the impact of skin type given the same side of the hand. From the right panel of Fig.~\subref*{fig:skin}, in the PD case, {only 0.04\% of the posterior distribution of the difference of means between lighter and darker skin groups} is within the ROPE, suggesting that the null hypothesis is rejected and the darker skin group outperforms the lighter skin group. In the PU case, {18\% of the posterior distribution falls within the ROPE. This observation is consistent with the participant-specific experiments that when using the back of the hand as the ROI, the skin color is more credible to be a factor in the accuracy of \spo2 estimation than using the palm.}

Second, we analyze the impact of the side of the hand for two skin color groups. For the darker skin group shown in the right panel of Fig.~\subref*{fig:skin}, {only 9\% of the posterior distribution of the difference of means of the PU and PD cases falls in the ROPE.} This shows that there is {a high} uncertainty in the estimate of zero difference, which is consistent with the results from the participant-specific experiments. 
However, unlike the participant-specific experiments, for the lighter skin group, {0.2\% of the posterior distribution of the difference of means between PU and PD cases {falls} in the ROPE. This suggests that the null hypothesis is rejected and that the PU outperforms the PD in the lighter skin group.} As for the mixed group illustrated in the right panel of Fig.~\subref*{fig:pd_pu}, {only 8\% of the posterior distribution of the difference of means {falls in} the ROPE,} suggesting that there is a high uncertainty to conclude that PU and PD cases are comparable. 

This different generalization capability in the PU and PD cases may be attributed to the skin color difference between the palm and the back of the hand.
The color of the back of the hand tends to be darker than the color of the palms and has larger color variation among participants due to different degrees of sunlight exposure. In contrast, the color variation of the palms is much milder among participants. 
Furthermore, in the participant-specific experiments, the individualized models learn the traits of the skin type and the side of the hand from each participant, whereas, in the leave-one-participant-out experiments, the learned model must capture the general characteristics of the population.}

\subsection{Ablation Studies}

To justify the use of nonlinear channel combinations and convolutional layers for temporal feature extraction in our proposed models, we conduct two ablation studies comparing the performance of these model components to other generic ones. 
We focus on the PU case to  avoid the uncontrolled impact of such factors as skin tone and hair. 
In the first ablation study, we compare nonlinear to linear channel combination. We create a variant of Model 1 with only a single linear channel combination layer with no activation function and repeat the leave-one-participant-out experiments.

In the second study, we compare the performance of using convolutional layers for temporal feature extraction to using fully-connected dense layers. We create this second variant of Model 1 
and repeat leave-one-participant-out experiments. 

\begin{table}[!t]
\caption{Numerical results of the ablation studies for Model~1~(M1) in the leave-one-participant-out mode. Comparisons among the proposed (nonlinear) M1, modified M1 with only linear channel combinations, and modified M1 with fully connected dense layers instead of convolutional layers are listed. Ablation studies confirm that the nonlinear channel combinations and convolutional layers improve model performance.}
\centering
\hspace{-1.5mm}
\resizebox{3.4in}{!}{%
\begin{tabular}{cc|ccc}
\hline
Method & & $\rho$ & MAE(\%) & RMSE(\%) \\ \hline \hline
{Linear Ch. Comb.} & Median &0.46 &  2.14&2.66  \\
{+ Conv. layer for Feat. Extra.} & IQR &0.38 &0.73 &0.93 \\ \hline
{Nonlinear Ch. Comb.} & Median & 0.41 & 2.29 & 2.66  \\
{+ Fully Connec. layer for Feat. Extra.}& IQR &0.39&0.63& 0.70\\ 
 \hline \hline
{Model 1 (Proposed): Nonlinear Ch. Comb.} & Median & \textbf{0.46} & \textbf{1.97} & \textbf{2.32} \\
 {+ Conv. layer for Feat. Extra.} & IQR & 0.36&0.80&0.87 \\ \hline
\end{tabular}
}
\label{tab:ablation table}
\end{table}

{TABLE~\ref{tab:ablation table} presents the medians and IQRs specified for numerical comparison of the ablation study. First, we compare the first and the third rows in TABLE~\ref{tab:ablation table} for ablation study 1. Our proposed Model~1 achieves a better correlation with a median of 0.46 and IQR of 0.36 and a better RMSE with a median of 2.32 and IQR of 0.87 than its linear channel combination variant. Besides, Model~1 achieves a comparable MAE with a better median of 1.97 but a wider IQR of 0.80. The overall better performance of Model~1 suggests the necessity of using the nonlinear channel combination method. Second, in ablation study 2, we compare the second and the third rows in TABLE~\ref{tab:ablation table}. We observe that Model~1 outperforms its second variant with fully connected layers for feature extraction with better medians in terms of correlation (0.46 vs. 0.41), MAE (1.97 vs. 2.29), and RMSE (2.32 vs. 2.66), and narrower IQR of correlation. This suggests that convolutional layers are better than fully connected layers for temporal feature extraction.}

\section{{Discussion}}

\subsection{Contact-based Dataset Testing}
{{We also test our models on the publicly available dataset gathered by Nemcova \emph{et al.} for their \spo2 estimation work~\cite{nemcova}. This dataset consists of contact-based {smartphone} video recordings where {a} participant {placed a} finger on the smartphone camera and {was} illuminated by the camera flashlight. Participants were asked to breathe normally {without following} any {sophisticated} breathing protocol. Each recording {lasts about 10 to 20} seconds{. The} subject for each recording is not identified, so subject-specific and leave-one-{participant-}out experiments {cannot be} conducted. There is a single reference \spo2 value associated with each recording. We used 14 recordings for training and {seven} recordings for testing and compared them with the modified {ratio-of-ratios} method proposed in their paper.}}

{
{{As shown in TABLE~\ref{tab:my-table}}, Models 1 and 2 {outperform} the method used by Nemcova~\emph{et al.} on both the training and test recordings. Model~3 {is} not able to generalize well from the training {set} to the test set, {which may be due to} the small size of the dataset. It should be noted that {because the participants were not asked to follow any sophisticated breathing protocol, the dynamic range of \spo2 values is narrow}. These results show that our CNN {Models 1 and 2} work well for contact-based video recordings in addition to contactless videos recordings.
}}

\begin{table}[!t]
\caption{Experimental results {of proposed methods} on the contact-based video \spo2 dataset {from}  Nemcova \emph{et al.}~\cite{nemcova}. One \spo2 estimate was {output} per recording and MAE and RMSE were calculated across all recordings. Models 1 and 2 outperform the method proposed by Nemcova \emph{et al.}, {Model} 3 was unable to generalize well to the test set.}
\centering

{\begin{tabular}{c|cc|cc}
\hline
 & \multicolumn{2}{c|}{MAE (\%)} & \multicolumn{2}{c}{RMSE (\%)} \\
 & Training & Test & Training & Test \\ \hline \hline
\multicolumn{1}{c|}{Model 1} & 0.86 & \textbf{1.19} & 0.94 & \textbf{1.36} \\
\multicolumn{1}{c|}{Model 2} & 0.50 & 1.28 & 0.59 & 1.64 \\
\multicolumn{1}{c|}{Model 3} & 0.75 & 3.28 & 0.99 & 3.69 \\ \hline \hline
\multicolumn{1}{c|}{Nemcova \emph{et al.}~\cite{nemcova}} & 2.05 & 2.18 & 2.24 & 2.36
\\
\hline
\end{tabular}%
}
\label{tab:my-table}
\end{table}

\subsection{{Ability to Track \spo2 Change}}

\begin{figure}[!t]
  \centering
  \includegraphics[width=3.5in]{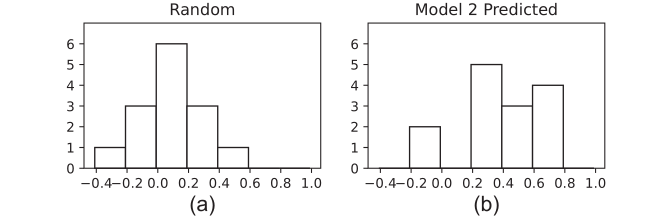}
  \caption{{{Histograms of correlation values} between reference \spo2 signals and (a)~randomly generated \spo2 signals, {or} (b)~\spo2 signals predicted by neural network Model 2. The correlation distribution for Model 2 is centered much higher than the random {guess, confirming Model~2's capability to track \spo2.}}}
  \label{fig:corr}
\end{figure}

{By employing the standard machine learning methodology of training-validation-test split in Section~\ref{experiment result} to {learn neural networks that perform well on unseen data}, we {have already ensured} the generalizability of our models ~\cite[Chapter~11]{Shwartz2014}. As further evidence that our models are capable of outputting meaningful predictions, we compare \spo2 predictions {from our learned models} to randomly generated \spo2 {values}. For each reference signal, a random prediction signal was generated by choosing \spo2 values between the minimum and maximum values from the reference signal and applying a moving average window in the same way as is applied to the neural network predictions. {Fig.~\ref{fig:corr}(a)} shows {a histogram} of the correlations between the reference \spo2 signals {and} the randomly generated predictions {and Fig.~\ref{fig:corr}(b) shows a histogram of correlations between the reference \spo2 signals and} the predictions generated by Model~2. It is {revealed that the neural network with a median correlation of $0.41$\footnote{{It has been shown in other applications that even low correlation coefficients can be meaningful. For example, in photo response non-uniformity (PRNU) work, the device used to take a photo can be predicted with correlation values below 0.1~\cite{baar2012camera}.}} {outperforms random guessing with a median correlation of $-0.02$, confirming Model~2's capability to track \spo2.}}}


\subsection{Visualizations of RGB Combination Weights}
 
\begin{figure}[!t]
  \centering

  \subfloat[]{\includegraphics[width=1.7in]{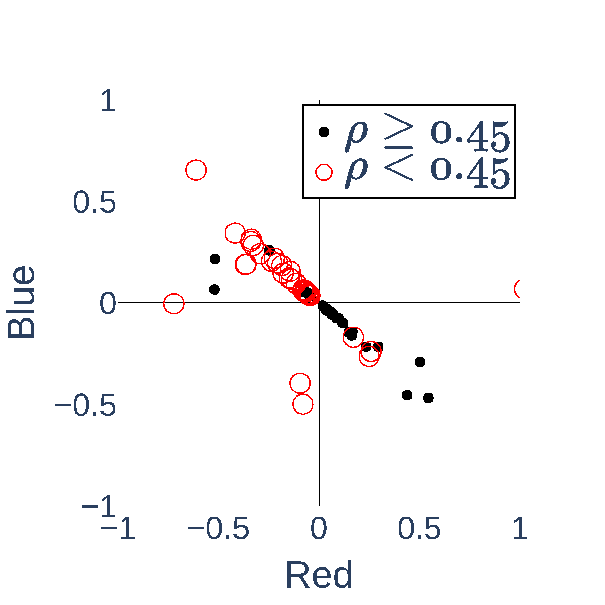}\label{fig:RB}}
  
  \subfloat[]{\includegraphics[width=1.7in]{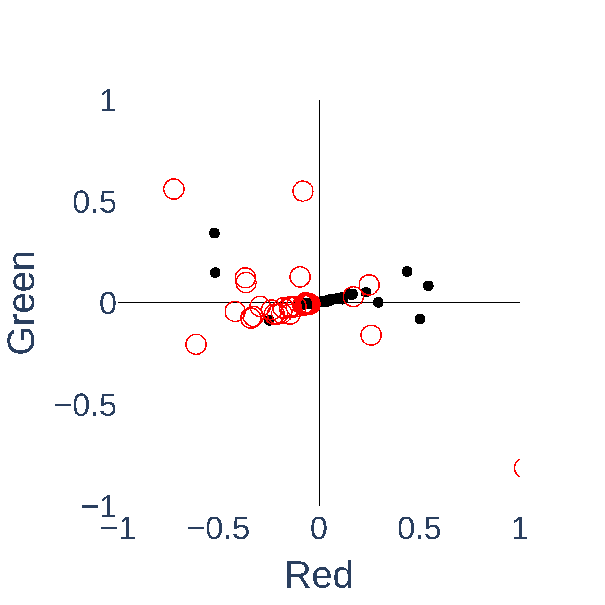}\label{fig:RG}}

  \hfill
  \subfloat[]{\includegraphics[width=1.7in]{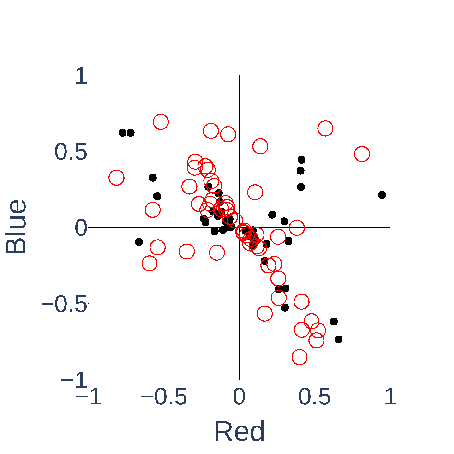}\label{fig:RB_7}}
  
  \subfloat[]{\includegraphics[width=1.7in]{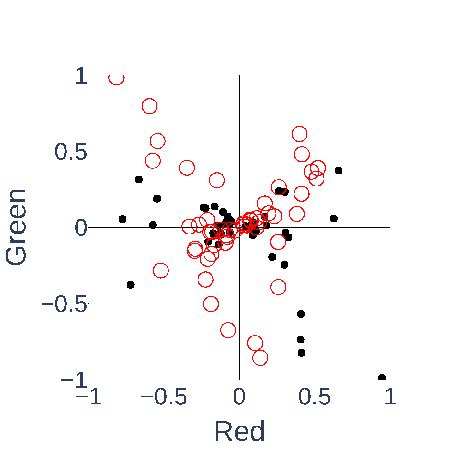}\label{fig:RG_7}}
  \caption{Learned RGB channel weights. Plots (a) and (b) are the channel weights learned by different model instances trained on the data of all study participants together, projected onto the RB and RG planes in the RGB space. Plots (c) and (d) are the RB and RG projections of the learned channel weights for model instances trained on random subsets of the participants' data. Each point is color-coded according to the correlation $\rho$ achieved by the instance.
  }

\end{figure}

To understand and explain what our physiologically inspired models have learned,
we conduct a separate investigation to visualize the learned weights for the RGB channels.
Our goal is to understand the best way to combine the RGB channels for \spo2{} prediction. 
Having an explainable model is important for a physiological prediction task like this. Our neural network models can be considered as nonlinear approximations of the hypothetically true function that can extract the physiological features related to \spo2{} buried in the RGB videos. The ratio-of-ratios method, for example, is another such extractor that combines the information from the different color channels at the end of the pipeline. For this experiment, we use the modified version of Model 1 from the ablation studies that has only a single linear channel combination at the beginning. Seeing that using a single linear channel combination did not significantly reduce model performance in the ablation studies, and understanding that the linear component may dominate the Taylor expansion of a nonlinear function, we use only linear combinations for this model to facilitate more interpretable visualizations.

We have trained 100 different instances of the model on the first two cycles from all the recordings and tested on the third cycle from all recordings. The difference between each instance is that the weights are randomly initialized. The weights for each channel learned by the model instances were visualized as points representing the heads of the linear combination vector in RGB space. Each point is colored according to the average test correlation achieved by the model instance. Figs.~\subref*{fig:RB}~and~\subref*{fig:RG} show the projections of these points onto the RB and RG planes. The subfigures reveal that the majority of the channel weights lay along certain lines in the RGB space. For the weights on the line, the ratio of the blue channel weight to the red channel weight is 0.87, the ratio of the green channel weight to red channel weight is 0.18. It is clear that the red and blue channels are the dominating factors for \spo2{} prediction.

To further verify this result, we repeat this experiment {under the following setup:} instead of using the data from all participants, for each model instance, we randomly select seven participants and use their data for training and testing. In this case, the difference between each model instance is not only the initialized weights but also the random subset of participants that the model was trained on. Fig.~\subref*{fig:RG_7} reveals that most of the better-performing instances (with $\rho \geq 0.45$)
have little contribution from the green channel. In Fig.~\subref*{fig:RB_7}, we again see that most of the points lay on a line in the RB plane, the ratio of the blue channel weight to the red channel weight for these points is 0.80.

These results are in accordance with the {bio}physical understanding of how light is absorbed by hemoglobin in the blood. {Recall that} Fig.~\ref{fig:extinction} reveals a large difference between the extinction coefficients, or the amount of light absorbed, by deoxygenated and oxygenated hemoglobin at the red wavelength. There is a significantly smaller difference at the blue wavelength and almost no difference at green. The amount of light absorbed influences the amount of light reflected which can be measured through the camera. A larger difference in extinction coefficients makes it easier to measure the ratio of light absorbed by oxygenated vs. deoxygenated hemoglobin over time. This ratio indicates the level of blood oxygen saturation.  Therefore, from a physiological perspective, it makes sense for the neural networks to give larger weight to the red and then blue channels and give little to the green channel. These visualizations indicate that the models are learning physically meaningful features.

\section{Conclusion {and Future Work}}
In this paper, we have proposed the first CNN-based work to solve the challenging problem of video-based remote \spo2{} estimation. We have designed three optophysiologically inspired neural network architectures. In both participant-specific and leave-one-participant-out experiments, our models are able to achieve better results than the state-of-the-art method. We have also analyzed the effect of skin color and the side of the hand on \spo2{} estimation and have found that in the leave-one-participant-out experiments, {the side of the hand} plays an important role, with better \spo2{} estimation results achieved in the palm-up case for the lighter-skin group. We have also shown the explainability of our designed architectures by visualizing the weights for the RGB channel combinations learned by the neural network, and have confirmed that the choice of the color band learned by the neural network is consistent with the established optophysiological methods. 

{In future work, we plan to investigate the impact of decreasing the  measurement {duration} since shorter measurement {duration} is generally preferred in clinical applications.
Another direction is to investigate the effectiveness and performance of the proposed \spo2{} estimation method in clinical or physiological applications while the participants are in motion.}

\balance
\bibliographystyle{IEEEtran}
\bibliography{ref}

\end{document}